\documentclass[twocolumn]{article}

\usepackage[utf8]{inputenc}
\usepackage[margin=0.7in]{geometry}
\usepackage{authblk}
\usepackage{setspace}
\usepackage{graphicx}
\graphicspath{{./figures/}}
\usepackage{subcaption}
\usepackage{amsmath}
\usepackage{booktabs}
\usepackage{hyperref}
\usepackage{multirow}
\usepackage{makecell}
\usepackage{xcolor}
\usepackage{enumitem}
\usepackage{needspace}
\usepackage[utf8]{inputenc}
\usepackage[T1]{fontenc}
\DeclareUnicodeCharacter{0308}{\"{}}
\usepackage{titlesec}

\titlespacing*{\section}
{0pt}
{1.5ex plus 0.4ex minus 0.2ex}
{0.6ex plus 0.2ex}

\titlespacing*{\subsection}
{0pt}
{1.2ex plus 0.3ex minus 0.2ex}
{0.6ex plus 0.2ex}

\newcommand{\etal}{\textit{et al}. }

\usepackage[switch]{lineno}       
\usepackage[
  style=nejm, 
  citestyle=numeric-comp,
  sorting=none
]{biblatex}
\title{\textbf{Social Pain Disrupts Emotion-Action Brain-State Dynamics in Adolescents with Non-Suicidal Self-Injury}}

\author[1,2]{Ying Xu}
\author[3]{Xiaojun Liang}
\author[1,2]{Li Zhang}
\author[4]{Yixuan Yuan}
\author[1,2]{Gan Huang}
\author[5*]{Yongjie Zhou}
\author[1,2*]{Zhen Liang}

\affil[1]{School of Biomedical Engineering, Medical School, Shenzhen University, Shenzhen, China.}
\affil[2]{Guangdong Provincial Key Laboratory of Biomedical Measurements and Ultrasound Imaging, Shenzhen, China.}
\affil[3]{Pengcheng Laboratory, Shenzhen, China.}
\affil[4]{Department of Electronic Engineering, The Chinese University of Hong Kong, Hong Kong, China.}
\affil[5]{Department of Child Health, Shenzhen Maternity and Child Healthcare Hospital, Shenzhen, China.}

\affil[*]{Address correspondence to: qingzhu1108@126.com and janezliang@szu.edu.cn}

\date{}

\begin{document}
\flushbottom
\twocolumn[
\begin{@twocolumnfalse}
\maketitle
\begin{abstract}
\textcolor{black}{Non-suicidal self-injury (NSSI) is highly prevalent among adolescents with depression, yet the rapid neurodynamic processes through which socially salient distress is translated into maladaptive behavioral tendencies remain poorly understood. Here, we integrate an experimental pain paradigm, electroencephalography (EEG) microstate analysis, and interpretable deep sequence modeling to characterize millisecond-scale brain-state dynamics associated with NSSI. EEG was recorded from 106 adolescents with depression, including 67 with NSSI (DN$+$) and 39 without NSSI (DN$-$), during social pain, physical pain, and resting-state conditions. A neurodynamic modeling framework is developed to capture higher-order dependencies within microstate sequences through joint disease-specific, domain-adversarial, consistency, and contrastive learning. Among the three conditions, social pain elicits the most discriminative NSSI-related neural dynamics. The proposed framework achieves a classification accuracy of 68.55\%, outperforming the best- performing baseline model by 8.94 percentage points. Importantly, model interpretation converges with conventionalmicrostate analyses to reveal disrupted bidirectional transitions between MS3 and MS5 in DN$+$ adolescents duringsocial pain. Source reconstruction links MS3 to emotional/interoceptive processing and MS5 to action preparation, indicating impaired emotion-action coupling. Time-resolved analyses further reveal enhanced recruitment of action-preparation states during early-to-middle stages of social pain processing, followed by increased engagement of emotion-related states at later stages in DN$+$ adolescents. Moreover, MS5-to-MS3 dynamics mediate social-evaluation sensitivity and affective outcomes in DN$-$ but not DN$+$, whereas MS3-to-MS5 transitions are associated with greater negative affect in DN+. Together, these findings identify disrupted emotion–action coupling as a key neurodynamic mechanism underlying altered social pain processing in adolescents with NSSI, providing a mechanistically interpretable neural signature for objective identification of NSSI.}
\end{abstract}

\begin{list}{}{\leftmargin=1.5em \rightmargin=1.5em}
\item\textbf{Keywords:} Non-suicidal self-injury; Social pain; EEG microstates; Emotion-action coupling; Neurodynamic modeling
\end{list}

\vspace{1em}
\end{@twocolumnfalse}
]

\section*{Introduction}
\textcolor{black}{Non-suicidal self-injury (NSSI) is a major mental health concern among adolescents, particularly those with depression. Large-scale epidemiological studies have estimated a lifetime prevalence of approximately 17.7\% in the general adolescent population, with substantially higher rates among adolescents with depression, more than half of whom may report a history of self-injurious behaviors\cite{muehlenkamp2012international,gillies2018prevalence}. Beyond its high prevalence, NSSI is associated with greater clinical severity and represents an important prospective risk factor for subsequent suicidal behavior\cite{klonsky2013relationship,ribeiro2016model}. However, the neurobiological processes through which distressing experiences are dynamically transformed into maladaptive behavioral tendencies remain poorly understood. Clinical identification of NSSI also continues to rely largely on self-report and clinical interviews, underscoring the need for objective neurophysiological measures that can complement conventional assessment\cite{hooley2020nonsuicidal}. Characterizing the neural dynamics associated with NSSI may therefore provide important mechanistic insights into this clinically heterogeneous behavior and facilitate its objective identification in adolescents with depression.}

\textcolor{black}{Pain processing provides a particularly relevant framework for understanding the neurobiology of NSSI. The Interpersonal Psychological Theory of Suicide (IPTS) emphasizes the roles of adverse interpersonal experiences and altered pain-related processes in the development of self-injurious and suicidal behaviors\cite{joiner2005people,van2010interpersonal}. Consistent with this framework, individuals with NSSI frequently exhibit atypical physical pain processing, including elevated pain thresholds and reduced subjective pain sensitivity\cite{koenig2017longitudinal}. Neuroimaging studies have further identified altered responses in pain-related regions, particularly the anterior cingulate cortex and insula, which contribute to the sensory, affective, and interoceptive dimensions of pain\cite{kross2011social,eisenberger2012pain,groschwitz2016differential}. Importantly, NSSI is also closely associated with adverse interpersonal experiences, making social pain particularly relevant to its neurobiology. Social rejection and physical pain engage partially overlapping neural systems\cite{eisenberger2003does}, suggesting that altered responses to social distress in NSSI may involve disrupted coordination among emotional, interoceptive, and action-related processes. Yet most previous studies have focused on regional activation or static neural measures. How the brain dynamically transitions between these functional processes as social distress unfolds, and whether such temporal organization is disrupted in adolescents with NSSI remains largely unknown.}

\textcolor{black}{Electroencephalography (EEG) microstates provide a powerful approach for addressing this question. EEG microstates are transient, quasi-stable scalp potential topographies that persist for tens to hundreds of milliseconds and are thought to represent rapidly changing functional configurations of large-scale brain networks\cite{lehmann1987eeg}. Unlike conventional time–frequency analyses that primarily characterize local spectral properties, microstate analysis captures the temporal organization of whole-brain electrophysiological activity, providing a systems-level representation of rapidly evolving brain states\cite{britz2010bold}. EEG microstates have been increasingly investigated in psychiatric and neurological disorders and provide a potential bridge between electrophysiological dynamics and large-scale functional brain networks\cite{lehmann2005eeg,da2020eeg,raeisi2025eeg,murphy2020abnormalities}. Altered microstate properties have also been reported in depression. Murphy \etal \cite{murphy2020abnormalities} identified abnormalities in microstate temporal properties in patients with major depressive disorder (MDD), whereas Zhao \etal \cite{zhao2022eeg} reported altered duration, occurrence, and coverage of multiple microstate classes in drug-naive adolescents with first-episode depression. Neuroimaging studies further suggest that canonical microstates are associated with intrinsic functional networks involved in salience processing, attention, sensory processing, and internally oriented cognition\cite{van2010eeg}. Moreover, abnormal microstate dynamics in MDD have been associated with cognitive dysfunction\cite{peng2024abnormalities}. Together, these findings suggest that microstate dynamics may provide a temporally resolved window into the large-scale neural alterations associated with psychopathology.}

\textcolor{black}{However, conventional microstate studies typically summarize brain dynamics using aggregate measures such as duration, occurrence, coverage, and pairwise transition probabilities. Although informative, these measures compress the temporal organization of microstate sequences and may therefore miss higher-order dependencies among rapidly evolving brain states. This limitation is particularly relevant to NSSI, in which pathological responses to socially salient distress may emerge not from abnormalities in a single brain state, but from disrupted transitions and coordination between states supporting emotional processing and behavioral preparation. Sequence-based machine learning provides a means to model these higher-order temporal dependencies directly\cite{khanna2015microstates}. More broadly, EEG-based machine-learning approaches have demonstrated the feasibility of identifying depression-related neural patterns from spectral, connectivity, and end-to-end representations\cite{gao2018machine,xia2023end}. Liang \etal \cite{liang2025nssi}, for example, developed a semi-supervised multi-concept framework for EEG-based identification of NSSI, whereas microstate-based approaches have achieved promising classification performance in MDD using conventional temporal parameters and sequence complexity\cite{li2021altered,zhao2022eeg}. Nevertheless, these approaches have predominantly treated microstate properties as aggregated features rather than modeling the microstate sequence itself. Consequently, the higher-order brain-state transitions that may distinguish adolescents with and without NSSI, particularly during socially salient distress, remain insufficiently characterized.}

\begin{figure*}[!t]
    \centering
    \includegraphics[width=\textwidth]{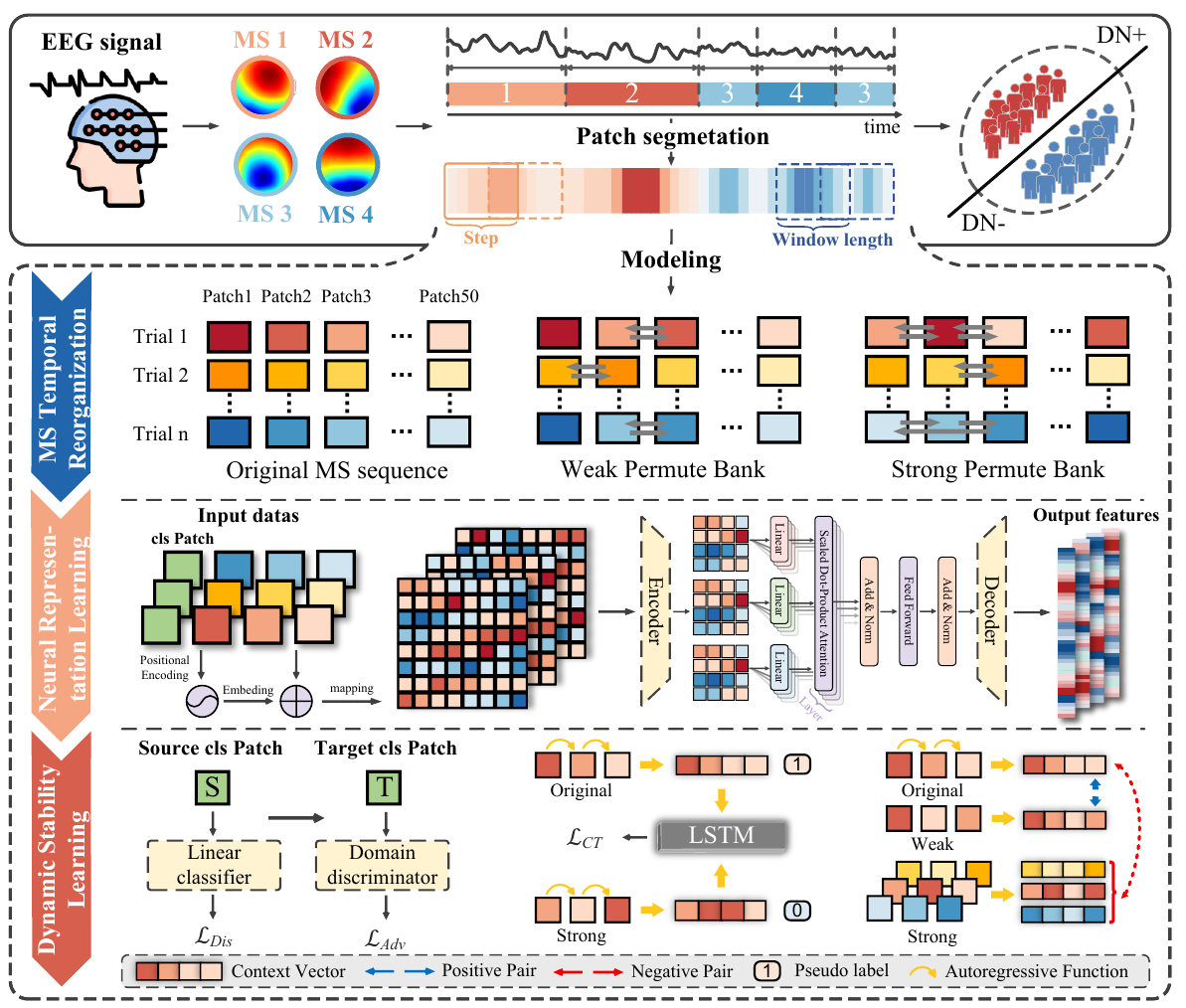}
    \caption{\textbf{Overview of the proposed model framework.} The model consists of four main components.(1) Microstate sequence construction. Preprocessed EEG signals are subjected to a two-step microstate clustering procedure using the k-means algorithm to obtain optimal microstate templates. These templates are then back-fitted to each data segment to generate corresponding microstate time series. (2) MS temporal reorganization. The microstate sequences are segmented into patches using sliding windows (window length = 50, step size = 50). Two shuffling strategies (weak permutation and strong permutation) are further applied to enhance the diversity of temporal patterns and improve model robustness. (3) Neural representation learning. A PatchTST architecture is employed to capture dynamic transition patterns within the microstate sequences and map them into high-dimensional representations. A dedicated CLS patch is introduced to facilitate the integration of local segment-level characteristics and global sequence-level structure. (4) Dynamic stability learning. The model is jointly optimized using multiple objectives, including a disease-specific loss, domain adversarial loss, consistency loss, and contrastive learning loss. This joint optimization enables the model to anchor disease-relevant pathological microstate transition patterns associated with NSSI, thereby samplifying subtle differences between patient subgroups.}
    \label{Fig: msppt step}
\end{figure*}

\textcolor{black}{Here, we integrate an experimental pain paradigm, EEG microstate analysis, and interpretable deep sequence modeling to investigate the neurodynamic mechanisms associated with NSSI in adolescents with depression (Fig.~\ref{Fig: msppt step}). We examine brain-state dynamics across social pain, physical pain, and resting-state conditions to determine which context most strongly exposes NSSI-related neural alterations. Rather than using microstates solely as static summary features, we directly model their sequential organization to capture higher-order transition dependencies and identify discriminative neurodynamic patterns. We further combine model interpretation with conventional microstate analysis, source-level reconstruction, functional decoding, time-resolved analysis, and mediation analysis to characterize the functional and behavioral relevance of the identified transitions. We hypothesize that social pain preferentially reveals disrupted transitions between brain states supporting emotional/interoceptive processing and action preparation in adolescents with NSSI. By linking millisecond-scale brain-state dynamics to individual-level identification, this framework aims to provide a mechanistically interpretable account of altered emotion–action coupling in NSSI.}

\section*{Results}
\textcolor{black}{To characterize the neurodynamic alterations associated with NSSI, we analyze EEG microstate sequences acquired from adolescents with depression during social pain, physical pain, and resting-state conditions. We first examine whether different pain-related contexts differentially exposed NSSI-related neural dynamics and then identify the microstate transitions contributing to these differences. Source-level reconstruction and functional decoding are subsequently used to characterize the neurobiological substrates of the identified transitions, while time-resolved and mediation analyses examine their temporal and behavioral relevance. Finally, we evaluate whether these neurodynamic patterns support individual-level identification of NSSI and assess the robustness and generalizability of the proposed modeling framework.}

\subsection*{Social pain preferentially reveals NSSI-related neurodynamic alterations}

\textcolor{black}{We first examine whether NSSI-related neurodynamic differences varied across social pain, physical pain, and resting-state conditions. Under identical evaluation settings, decoding performance shows a consistent condition-dependent gradient, with social pain yielding the highest performance (Accuracy = 68.55\%, Precision = 69.49\%, Recall = 67.37\%, F1-score = 68.08\%, AUC = 68.06\%), followed by physical pain (Accuracy = 66.45\%) and resting state (Accuracy = 61.84\%) (Table~\ref{tab: state compare}). These results indicate that NSSI-related neural differences are most discriminative during social pain processing.}

\textcolor{black}{This pattern is consistent with the high self-relevance and affective salience of socially adverse experiences, which engage neural systems involved in emotional processing, interoception, and behavioral regulation\cite{eisenberger2003does,kross2011social,rotge2015meta}. By comparison, physical pain more directly engages nociceptive and somatosensory processes\cite{apkarian2005human,tracey2007cerebral}, whereas the resting state lacks an explicit affective challenge\cite{biswal1995functional,raichle2001default}. Thus, rather than indicating a generalized alteration across brain states, the observed performance gradient suggests that socially salient distress preferentially exposes neurodynamic differences associated with NSSI.}

\begin{figure*}[!t]
    \centering
    \includegraphics[width=\textwidth]{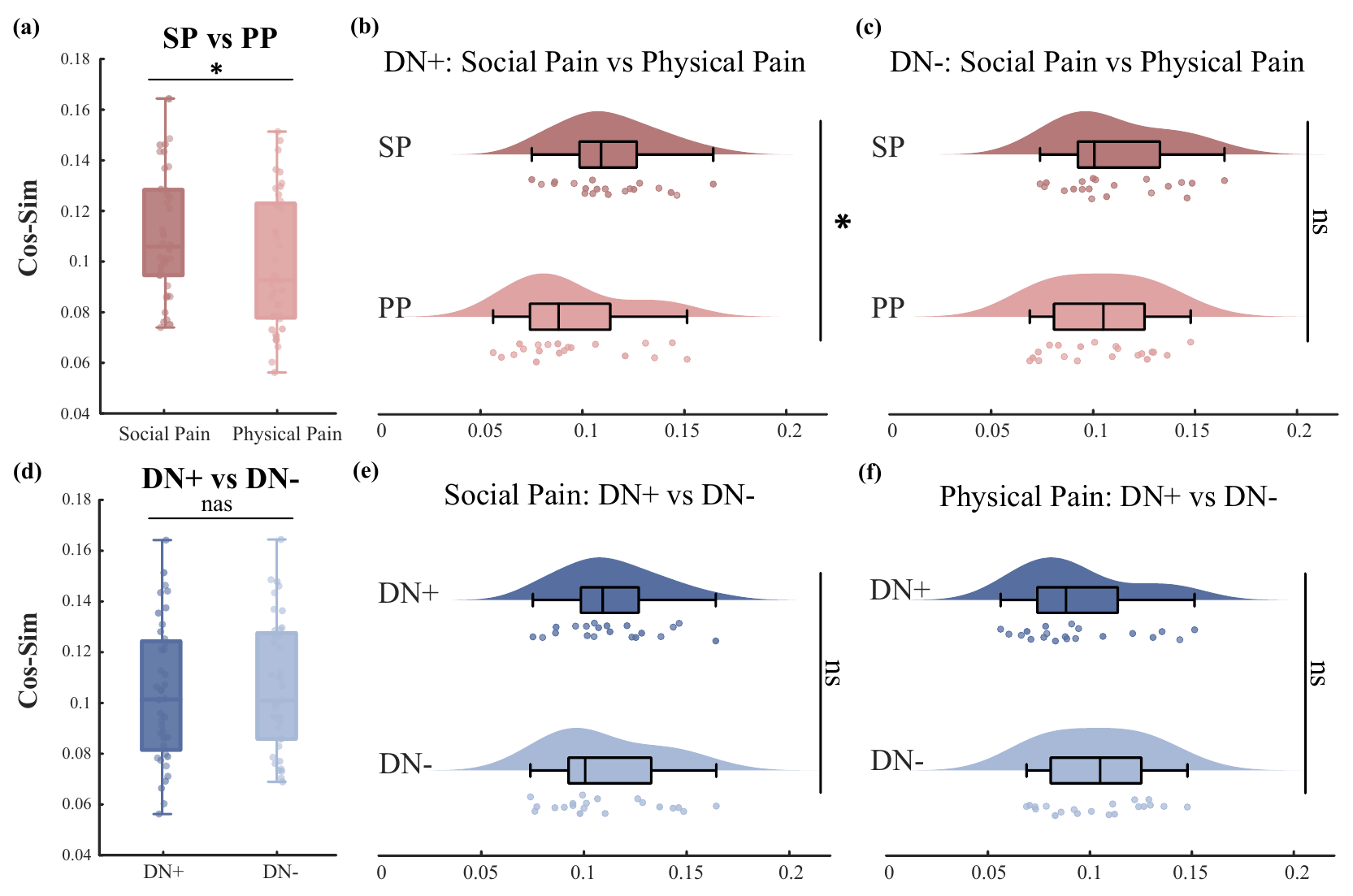}
    \caption{\textbf{Cosine similarity between model-derived EEG representations and stimulus representations across pain conditions.} (a) Group-level comparison of cosine similarity between model-derived EEG features and stimulus image features under social pain (SP) and physical pain (PP) conditions. (b-c) Within-group comparisons of SP versus PP for DN$+$ and DN$-$ adolescents, respectively. (d) Between-group comparison of cosine similarity across conditions. (e-f) Post hoc comparisons between DN$+$ and DN$-$ groups under SP and PP conditions, respectively. *p$<$0.05.}
    \label{Fig: cossim}
\end{figure*}

\textcolor{black}{To further examine condition-dependent neural-stimulus correspondence, we use CLIP\cite{radford2021learning} to extract visual representations of the social and physical pain stimuli and quantify their cosine similarity with model-derived EEG representations. A two-way mixed ANOVA is performed with pain type as the within-subject factor and group as the between-subject factor. As shown in Fig.~\ref{Fig: cossim}, EEG-stimulus representational similarity differs between social and physical pain conditions in DN+ (p = 0.02, t = 2.38). No significant between-group differences are detected within either pain condition, indicating that the effect is primarily condition dependent rather than reflecting a global difference in representational similarity between DN$+$ and DN$-$ adolescents. Together with the decoding results, these findings further support social pain as a particularly informative context for revealing NSSI-related neurodynamic alterations.}

\begin{table*}[!t]
    \caption{Model performance across social pain, physical pain, and resting-state conditions}    
    \centering
    \begin{tabular*}{\hsize}{@{}@{\extracolsep{\fill}}cccccc@{}}
        \toprule
        State & Accuracy(\%) & Precision(\%) & Recall(\%) & F1-score(\%) & AUC(\%) \\
        \midrule
        Resting & 61.84$\pm$01.39 & 62.26$\pm$03.10 & 62.11$\pm$10.00 & 61.60$\pm$04.13 & 60.50$\pm$02.64 \\
        Physical Pain & 66.45$\pm$04.31 & 67.07$\pm$05.49 & 66.58$\pm$11.31 & 66.12$\pm$05.98 & 66.00$\pm$05.73 \\
        Social Pain & \textbf{68.55$\pm$03.07} & \textbf{69.49$\pm$04.60} & \textbf{67.37$\pm$06.70} & \textbf{68.08$\pm$03.41} & \textbf{68.06$\pm$04.57} \\
        \bottomrule
    \end{tabular*}
    \label{tab: state compare}
\end{table*}

\subsection*{Disrupted MS3-MS5 transitions characterize social pain processing in NSSI}

Given that social pain most strongly differentiated DN$+$ from DN$-$ adolescents, we next investigate the neurodynamic patterns underlying this effect. Directed temporal dependencies between consecutive microstates are quantified from the attention weights of the trained model and aggregate into microstate transition attention matrices. Group-specific matrices are then examined to identify transitions that contributed most strongly to the discrimination of NSSI status.

\begin{figure*}[!t]
    \centering
    \includegraphics[width=\textwidth]{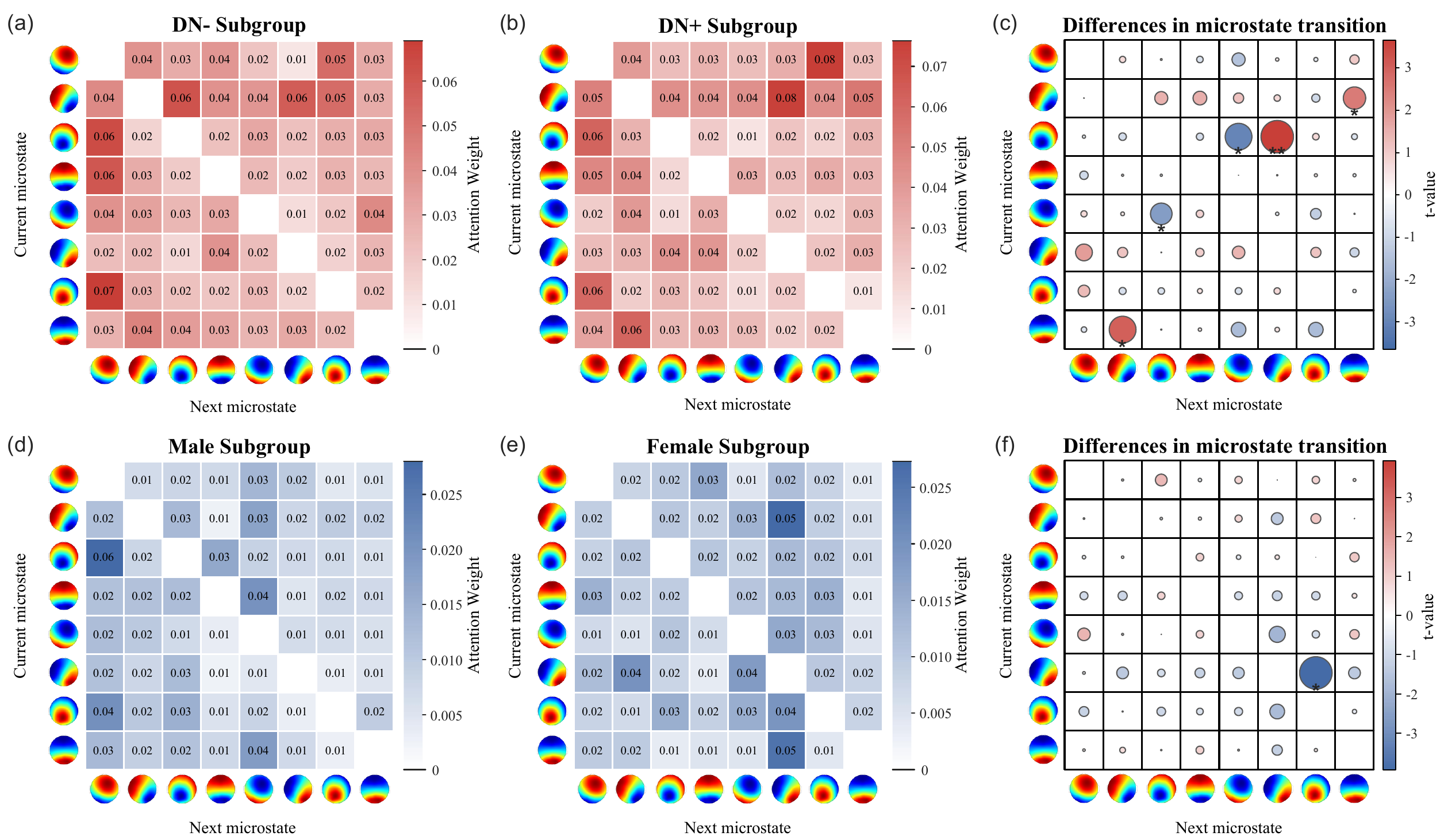}
    \caption{\textbf{Microstate transition attention patterns across clinical and sex subgroups.} (a-b) Mean microstate transition attention matrices for DN$+$ and DN$-$ adolescents, averaged across ten-fold cross-validation. (c) Statistical comparison of transition attention weights between clinical groups. (d-e) Mean transition attention matrices for male and female subgroups. (f) Statistical comparison between sex subgroups. Asterisks indicate transitions showing significant group differences. *p$<$0.05.}
    \label{Fig: attn heatmap}
\end{figure*}

During social pain processing, group differences are observed primarily for transitions involving MS2$\leftrightarrow$MS8, MS3$\leftrightarrow$MS5, and MS3$\rightarrow$MS6 (Fig.~\ref{Fig: attn heatmap}). Among these transitions, MS3$\leftrightarrow$MS5 shows the most consistent group difference. Importantly, this pattern converges with conventional microstate analyses based on occurrence, duration, coverage, and transition probability, which independently implicate MS3 and MS5 in differentiating DN$+$ from DN$-$ adolescents. DN$+$ adolescents exhibit weaker bidirectional MS3$\leftrightarrow$MS5 transition strength than DN$-$ adolescents, indicating attenuated dynamic coupling between these two brain states during social pain processing.

We additionally examine whether these transition patterns are strongly influenced by sex. In contrast to the clinical-group effects, sex-related differences are limited, with a significant difference observed only for the MS6$\rightarrow$MS7 transition. Details of the statistical analysis can be found in the supplementary materials Appendix I. These findings suggest that the prominent MS3$\leftrightarrow$MS5 alteration is more closely associated with NSSI status than with sex-related variation.

\subsection*{MS3-MS5 dynamics link emotional-interoceptive processing to action preparation}

\begin{figure*}[!t]
    \centering
    \includegraphics[width=\textwidth]{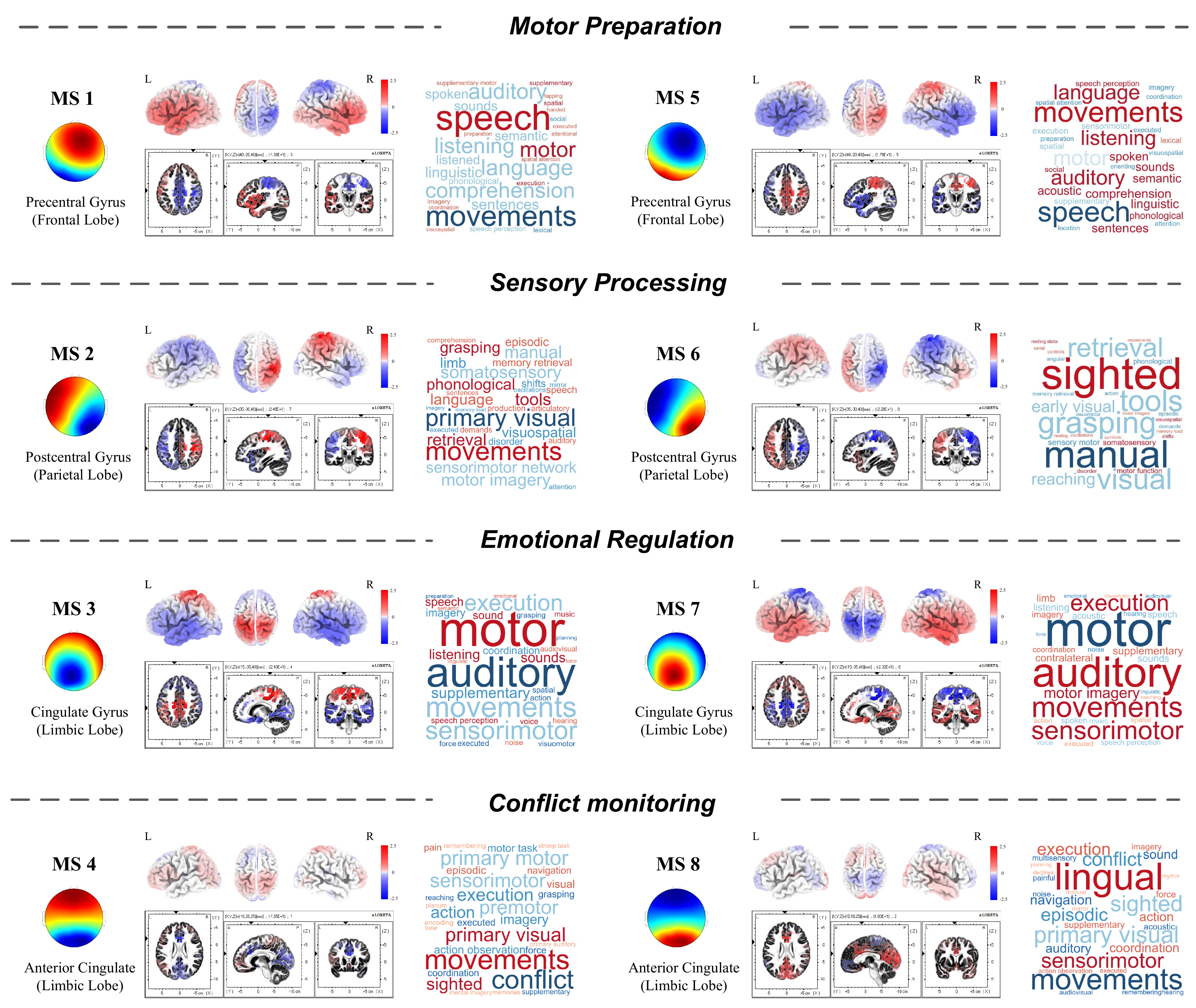}
    \caption{\textbf{Source-level reconstruction and functional annotation of EEG microstates.} Source-level activation patterns of the eight microstate templates are shown together with their meta-analytic functional annotations derived from Neurosynth. For each template, word clouds display the top 15 positively and negatively associated functional terms based on correlation coefficients, excluding anatomical terms. The cortical region showing the strongest source activity for each template is indicated below the corresponding map.}
    \label{Fig: neurosynth}
\end{figure*}

To characterize the functional significance of the disrupted MS3$\leftrightarrow$MS5 dynamics, we project microstate topographies into source space using the TESS framework\cite{pizzagalli2007electroencephalography} and perform meta-analytic functional decoding using Neurosynth. Source reconstruction reveal partially distinct spatial distributions for MS3 and MS5 (Fig.~\ref{Fig: neurosynth}). MS3 is predominantly associated with cingulate and medial frontal regions implicated in emotional and interoceptive processing, whereas MS5 shows stronger involvement of the precentral gyrus and adjacent premotor regions associated with action preparation.

These spatial and functional profiles suggest that MS3$\leftrightarrow$MS5 transitions index dynamic coordination between emotional/interoceptive processing and action-preparatory systems. The reduce bidirectional MS3$\leftrightarrow$MS5 transitions observed in DN$+$ adolescents therefore point to disrupted emotion-action coupling during the processing of socially salient distress. Rather than reflecting an abnormality confined to a single neural state, NSSI is characterized by altered coordination between brain states involved in evaluating affective information and preparing behavioral responses.

\subsection*{Time-resolved dynamics reveal an early action-preparation bias followed by late emotional recruitment}

\begin{figure*}[!t]
    \centering
    \includegraphics[width=\textwidth]{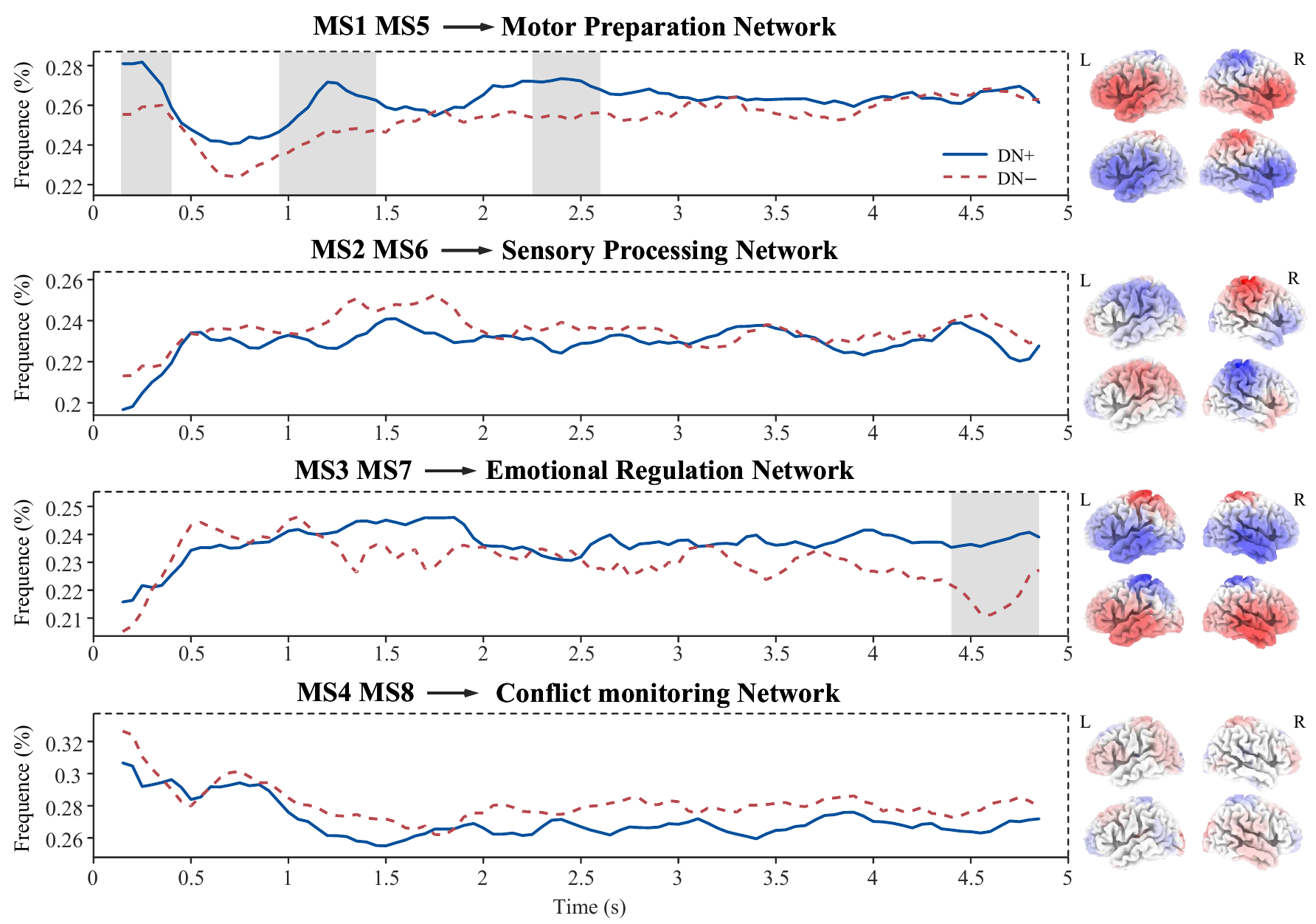}
    \caption{\textbf{Time-resolved validation of microstate dynamics during social pain processing.} A sliding-window approach (window length = 300 ms; step size = 50 ms) is used to quantify the temporal occurrence of each microstate. Microstate templates with identical spatial topographies but opposite polarity were treated as the same state. Solid lines represent adolescents with depression and NSSI (DN$+$), whereas dashed lines represent adolescents with depression without NSSI (DN$-$). Shaded regions indicate time intervals showing significant between-group differences. Corresponding source-level activation patterns are displayed for each microstate.}
    \label{Fig: temporal effect}
\end{figure*}

To determine how the identified emotion--action alterations unfold over time, we perform a time-resolved analysis of microstate dynamics during social pain processing. EEG responses are segmented using a sliding window of 300 ms with a 50-ms step, and the proportion of each microstate within each window is quantified. Microstates with identical spatial topographies but opposite polarities are treated as the same state. Individual-level estimates are subsequently averaged to generate time-resolved state-occurrence curves for DN$+$ and DN$-$ adolescents (Fig.~\ref{Fig: temporal effect}).

Distinct temporal differences emerge between groups. During the early-to-middle stages of social pain processing (0--400 ms, 950--1450 ms, and 2250--2600 ms), DN$+$ adolescents show greater recruitment of the state associated with motor and action preparation than DN$-$ adolescents. In contrast, during a later interval (4400--4850 ms), DN$+$ adolescents show greater recruitment of the state associated with emotional processing.

These results reveal a temporally structured alteration in social pain processing in NSSI, characterized by enhanced engagement of action-preparatory systems during early-to-middle processing followed by increased recruitment of emotion-related systems at later stages. This temporal organization is consistent with the abnormal MS3$\leftrightarrow$MS5 transitions identified above and provides complementary evidence that altered emotion-action coupling in NSSI unfolds dynamically over the course of social pain processing\cite{glenn2009social,menon2011large,kaiser2015large}.

\subsection*{MS5-to-MS3 dynamics differentially link social evaluation to affective outcomes}

We next examine whether the identified emotion-action dynamics are associated with individual differences in sensitivity to the social environment. Given the established relevance of peer interaction and social evaluation to adolescent NSSI\cite{nock2009people,prinstein2001adolescents}, mediation analyses are performed separately in DN$-$ and DN$+$ adolescents. Environmental variables include parent-child relationship factors, academic stress, and sensitivity to teachers' and peers' evaluations, while clinical and affective outcomes include PHQ-9\cite{spitzer1999validation}, GAD-7\cite{spitzer2006brief}, and subjective negative affect following social pain.

In DN$-$ adolescents, sensitivity to social evaluation is positively associated with MS5$\rightarrow$MS3 transition strength ($a=0.451$). Greater MS5$\rightarrow$MS3 transition strength is, in turn, negatively associated with PHQ-9 scores ($b=-0.480$) and subjective emotional ratings ($b=-0.493$). Bootstrap mediation analyses showed significant indirect associations between social-evaluation sensitivity and PHQ-9 scores (indirect effect $=-0.206$), as well as between social-evaluation sensitivity and subjective emotional ratings (indirect effect $=-0.232$). The corresponding direct effects remained significant, consistent with partial statistical mediation.

This pattern is absent in DN$+$ adolescents. Although sensitivity to social evaluation show direct associations with PHQ-9 scores and subjective emotional ratings, its association with MS5$\rightarrow$MS3 transition strength is not significant ($a=0.082$), nor are the associations between MS5$\rightarrow$MS3 dynamics and PHQ-9 ($b=-0.090$) or subjective emotional ratings ($b=-0.140$). Consequently, no significant indirect effects through MS5$\rightarrow$MS3 are observed.

Notably, the reverse MS3$\rightarrow$MS5 transition shows a significant association with subjective negative affect in DN$+$ adolescents. Together, these findings reveal a direction-dependent alteration in emotion-action brain-state dynamics: whereas action-to-emotion transitions are associated with social-evaluation sensitivity and affective outcomes in adolescents without NSSI, this relationship is disrupted in adolescents with NSSI, in whom stronger emotion-to-action dynamics are instead associated with greater negative affect.

\begin{figure*}[!t]
    \centering
    \includegraphics[width=\textwidth]{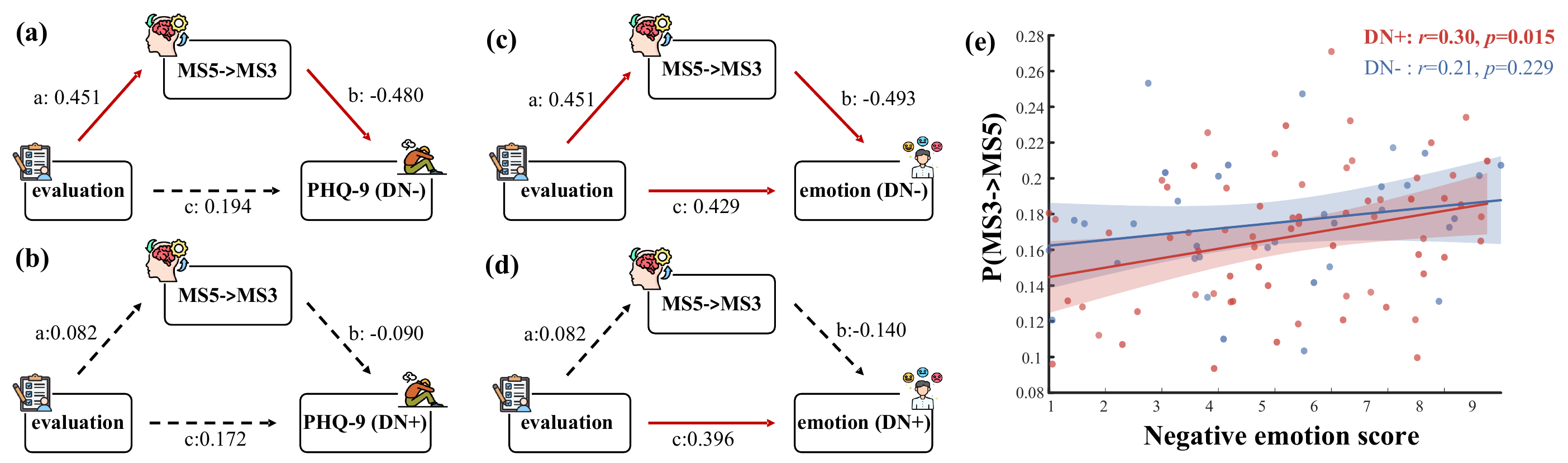}
    \caption{\textbf{Associations among social evaluation, microstate dynamics, and affective outcomes.} (a-b) Statistical mediation analyses examining MS5$\rightarrow$MS3 dynamics in the association between sensitivity to social evaluation and depressive symptoms (PHQ-9 scores) in DN$-$ and DN$+$ adolescents. (c-d) Corresponding analyses for subjective negative emotional ratings following pain-related stimuli. (e) Association between MS3$\rightarrow$MS5 transition probability and subjective negative emotional ratings across groups. Solid red lines indicate statistically significant paths, whereas dashed black lines indicate non-significant paths. Path coefficients are shown alongside each path. DN$+$, depression with NSSI; DN$-$, depression without NSSI.}
    \label{Fig: intermedia}
\end{figure*}

\subsection*{Neurodynamic modeling enables individual -level identification of NSSI}

Having identified disrupted emotion--action dynamics during social pain, we next evaluate whether the sequential organization of microstates provides sufficient information for individual-level identification of NSSI. The proposed model is compared with four conventional machine-learning methods\cite{cover1967nearest,breiman1984classification,cortes1995support,breiman2001random}, four neural sequence models\cite{rumelhart1986learning,hochreiter1997long,cho2014learning,vaswani2017attention}, five domain-adaptation approaches\cite{tzeng2014deep,long2015learning,ganin2016domain,sun2016deep}, and two contrastive-learning methods\cite{he2020momentum,chen2020simple} using identical inputs (Table~\ref{tab:NSSI model results}).

Conventional machine-learning methods show relatively limited discrimination, with accuracies ranging from 46.45\% to 53.55\%. Neural sequence models yield modest improvements, with the Transformer achieving the highest accuracy among the baseline methods (59.61\%). Domain-adaptation and contrastive-learning approaches produce comparable performance, with DDC reaching a precision of 65.17\% and SimCLR an accuracy of 59.47\%. In comparison, the proposed neurodynamic model achieves an accuracy of 68.55\%, precision of 69.49\%, recall of 67.37\%, F1-score of 68.08\%, and AUC of 68.06\%. Its accuracy exceeds that of the best-performing baseline model by 8.94 percentage points, indicating that explicitly modeling higher-order dependencies within microstate sequences improves the identification of NSSI-related neurodynamic patterns.

\begin{table*}[!t]
    \caption{Performance comparison with conventional machine-learning, neural sequence, domain-adaptation, and contrastive-learning methods.} 
    \centering
    \begin{tabular*}{\hsize}{@{}@{\extracolsep{\fill}}cccccc@{}}
        \toprule
        Method & Accuracy(\%) & Precision(\%) & Recall(\%) & F1-score(\%) & AUC(\%) \\
        \midrule
        KNN\cite{cover1967nearest} & 53.55$\pm$04.47 & 53.15$\pm$03.87 & \underline{61.58$\pm$07.87} & \underline{56.86$\pm$04.77} & 47.47$\pm$05.78 \\
        Decision tree\cite{breiman1984classification} & 47.50$\pm$04.62 & 47.30$\pm$04.60 & 49.21$\pm$11.84 & 47.84$\pm$07.62 & 48.33$\pm$04.00\\
        SVM\cite{cortes1995support} & 46.45$\pm$04.96 & 46.34$\pm$05.25 & 46.84$\pm$08.02 & 46.48$\pm$06.23 & 55.37$\pm$06.17 \\
        Random forest\cite{breiman2001random} & 52.24$\pm$06.30 & 52.23$\pm$06.07 & 52.89$\pm$08.08 & 52.44$\pm$06.63 & 52.47$\pm$06.91\\
        \midrule
        RNN\cite{rumelhart1986learning} & 58.03$\pm$05.80 & 64.11$\pm$10.92 & 49.74$\pm$26.82 & 50.72$\pm$15.13 & 56.11$\pm$09.70 \\
        LSTM\cite{hochreiter1997long} & 56.45$\pm$04.09 & 65.08$\pm$10.70 & 38.95$\pm$24.46 & 44.27$\pm$11.79 & 53.36$\pm$05.40 \\
        GRU\cite{cho2014learning} & 56.18$\pm$03.46 & 64.96$\pm$10.27 & 35.53$\pm$21.39 & 42.13$\pm$11.89 & 52.91$\pm$05.05 \\
        Transformer\cite{vaswani2017attention} & \underline{59.61$\pm$06.02} & 60.13$\pm$25.02 & 50.79$\pm$31.51 & 49.79$\pm$24.57 & 54.92$\pm$13.01 \\
        \midrule
        DDC\cite{tzeng2014deep} & 57.24$\pm$03.63 & \underline{65.17$\pm$10.66} & 47.89$\pm$27.29 & 49.54$\pm$12.26 & \underline{56.48$\pm$04.82} \\
        DAN\cite{long2015learning} & 56.18$\pm$04.07 & 55.42$\pm$21.78 & 44.21$\pm$31.18 & 44.16$\pm$22.20 & 53.42$\pm$06.75 \\
        DANN\cite{ganin2016domain} & 55.13$\pm$04.58 & 58.50$\pm$09.68 & 59.74$\pm$25.49 & 55.01$\pm$09.30 & 55.78$\pm$06.64 \\
        CORAL\cite{sun2016deep} & 48.68$\pm$05.34 & 48.66$\pm$05.02 & 46.84$\pm$09.35 & 47.39$\pm$06.84 & 48.20$\pm$03.39 \\
        DCORAL\cite{sun2016deep} & 56.05$\pm$03.78 & 60.03$\pm$25.64 & 36.32$\pm$24.24 & 40.78$\pm$18.85 & 52.70$\pm$07.93 \\
        \midrule
        MoCo\cite{he2020momentum} & 55.53$\pm$05.26 & 61.15$\pm$09.77 & 35.79$\pm$08.61 & 44.24$\pm$05.91 & 53.97$\pm$07.97 \\
        SimCLR\cite{chen2020simple} & 59.47$\pm$03.50 & 64.31$\pm$13.00 & 51.84$\pm$14.99 & 54.94$\pm$09.24 & 54.76$\pm$06.26 \\
        \midrule
        \textbf{Our model} & \textbf{68.55$\pm$03.07} & \textbf{69.49$\pm$04.60} & \textbf{67.37$\pm$06.70} & \textbf{68.08$\pm$03.41} & \textbf{68.06$\pm$04.57} \\
        \bottomrule
    \end{tabular*}
    \label{tab:NSSI model results}
\end{table*}

\section*{Discussion}
\textcolor{black}{The present study suggests that NSSI in adolescents with depression is characterized by altered coordination between emotional/interoceptive and action-related brain states during socially salient distress. These findings extend previous accounts of NSSI from abnormalities in individual regions or networks toward a dynamic systems perspective, in which vulnerability may arise from how rapidly evolving neural states interact over time.}

\subsection*{Social pain as a context for revealing NSSI-related vulnerability}

Interpersonal experiences are particularly salient during adolescence, when sensitivity to peer acceptance, rejection, and social evaluation increases substantially. This developmental context may help explain why socially painful experiences provide an informative window into NSSI-related neural dysfunction. The Interpersonal Psychological Theory of Suicide emphasizes adverse interpersonal experiences and altered pain-related processes in self-injurious and suicidal behaviors\cite{joiner2005people,van2010interpersonal}. Although NSSI is distinct from suicidal behavior, this framework highlights the potential importance of interactions between interpersonal distress and pain-related processing.

Physical and social pain recruit partially overlapping neural systems involved in affective salience and interoception\cite{kross2011social,eisenberger2012pain,groschwitz2016differential}. Social pain, however, additionally carries information about rejection, belonging, and self-relevance, placing greater demands on emotional appraisal and behavioral regulation. Social pain may therefore provide an experimental context that amplifies underlying vulnerabilities in processing socially salient distress, rather than representing a specific causal factor for NSSI.

\subsection*{Emotion-action decoupling as a neurodynamic mechanism of NSSI}

A central implication of the present findings is that NSSI-related dysfunction may involve impaired coordination between systems supporting emotional/interoceptive processing and those supporting action preparation. The cingulate cortex occupies an important position at the interface of affect, interoception, cognitive control, and action\cite{craig2009you,shackman2011integration,rotge2015meta}, whereas precentral and premotor systems contribute to the preparation and organization of behavioral responses. Flexible interaction between these systems may therefore be necessary for translating emotional information into context-appropriate behavior.

Previous neuroimaging studies of adolescent NSSI and depression have identified abnormalities across salience, default-mode, attention, and sensorimotor networks\cite{hu2023resting,tse2024mega}. Our findings extend these observations by suggesting that the relevant dysfunction may lie not only within individual systems but also in their temporal coordination. Inefficient switching between emotional and action-related states could impair the flexible transformation of socially salient information into adaptive behavioral responses.

This dynamic perspective may be particularly relevant to NSSI, which has frequently been conceptualized as a maladaptive strategy for regulating intense negative affect\cite{groschwitz2012neurobiology}. The observed temporal organization further suggests that action-related processing and emotional processing are abnormally coordinated as social distress unfolds. Rather than indicating a direct neural pathway to self-injury, these findings point to altered emotion--action integration as a broader mechanism through which distress may become more closely linked to behavioral readiness.

\subsection*{Directionality may distinguish emotional integration from behavioral readiness}

The directional organization of brain-state transitions suggests that emotion-action coupling is not a unitary process. Transitions from action-related toward emotional/interoceptive states may reflect incorporation of externally oriented information into internal affective appraisal, whereas transitions in the opposite direction may reflect transformation of emotional states toward behavioral preparation.

Such reciprocal coordination is important for adaptive emotion regulation, which requires continuous updating of internal states according to environmental demands and flexible selection of behavioral responses\cite{etkin2015neural,dalgleish2017social}. Disruption of this exchange could therefore reduce the flexibility with which socially salient experiences are processed.

Within this framework, the distinct behavioral associations of the two transition directions suggest an altered balance between emotional integration and action readiness in NSSI. Importantly, this interpretation does not imply that an individual transition directly represents self-injurious behavior. Rather, it suggests that negative affect may become more closely coupled to action-oriented neural configurations, providing a potential neurodynamic bridge between emotional dysregulation and maladaptive behavioral tendencies.

\subsection*{Multidimensional analyses support model robustness}

Beyond its mechanistic interpretation, the proposed framework showed stable performance across complementary validation analyses. The ablation analysis (Supplementary Appendix II) indicates that the disease-specific, domain-adversarial, consistency, and contrastive objectives make complementary contributions to the learned representation. In particular, the performance reductions observed after removing individual objectives suggest that the model benefits from jointly preserving disease-relevant information, reducing inter-individual domain variation, and maintaining discriminative temporal structure.

\textcolor{black}{Additional supplementary analyses further support the reliability of the proposed framework. The loss curves of the individual optimization objectives exhibit stable convergence throughout training, indicating that the multi-objective optimization strategy does not introduce conflicting optimization dynamics (Supplementary Appendix III). Permutation testing confirms that the observed classification performance remains significantly above chance level ($p=0.03$), while repeated experiments under different random sampling strategies demonstrate the robustness of the framework to sampling variability (Supplementary Appendix IV). Furthermore, comparisons of different PatchTST classification heads consistently favor the proposed design, suggesting that the performance improvement arises from the overall neurofunctional framework rather than architectural choices alone (Supplementary Appendix V).}

\textcolor{black}{Hyperparameter analyses further demonstrate the stability of the proposed framework. Across different loss weighting schemes and permutation ratios, the model maintains consistently competitive performance, with the optimal configuration achieving the best balance between representation learning and classification accuracy (Supplementary Appendix VI). These findings indicate that the framework is not overly sensitive to hyperparameter selection, supporting its robustness and potential for future clinical translation.}

\subsection*{Sex-related stability of neurodynamic representations}
Given the sex imbalance commonly observed in adolescent NSSI cohorts, we further examine whether the learned representations are strongly dependent on sex. As summarized in Table~\ref{tab: gender compare} and supplementary Appendix VII, within-sex and cross-sex analyses show broadly comparable discrimination, suggesting that the major neurodynamic information captured by the framework is not restricted to one sex. Under gender-consistent settings, performance decreases slightly but remains stable, with Accuracy/F1-score of 67.63\%/66.87\% in the female subgroup and 65.79\%/65.00\% in the male subgroup. In cross-gender evaluations, performance declines further, with relatively balanced results in the Female-to-Male condition (F1-score: 65.30\%), while the Male-to-Female condition shows high Precision (72.51\%) but markedly reduced Recall (53.68\%), leading to a lower F1-score (61.37\%). Overall, the model maintains good discriminative ability across single-gender and cross-gender settings, effectively distinguishing DN$+$ and DN$-$ patients.

\begin{table*}[!t]
    \centering
    \caption{Sex-specific and cross-sex validation of the proposed framework.}
    \begin{tabular*}{\hsize}{@{}@{\extracolsep{\fill}}cccccc@{}}
        \toprule
         Train set &  Test set & Accuracy(\%) & Precision(\%) & Recall(\%) & F1-score(\%) \\
         \midrule
         Entire data & Entire data & 68.55$\pm$03.07 & 69.49$\pm$04.60 & 67.37$\pm$06.70 & 68.08$\pm$03.41 \\
         Female subgroup & Female subgroup & 67.63$\pm$06.62 & 67.76$\pm$04.90 & 67.89$\pm$17.59 & 66.87$\pm$09.82 \\
         Female subgroup & Male subgroup & 65.79$\pm$02.79 & 66.48$\pm$04.36 & 65.26$\pm$10.43 & 65.30$\pm$04.74 \\
         Male subgroup & Male subgroup & 65.79$\pm$02.46 & 67.35$\pm$06.00 & 65.79$\pm$16.95 & 65.00$\pm$06.41 \\
         Male subgroup & Female subgroup & 66.32$\pm$02.00 & 72.51$\pm$05.64 & 53.68$\pm$05.46 & 61.37$\pm$02.23 \\
         \bottomrule
    \end{tabular*}
    \label{tab: gender compare}
\end{table*}

\subsection*{Independent datasets distinguish generalizability from disease specificity}
To further validate the decoding capability of the proposed model, we conduct additional evaluations on two independent public EEG datasets of MDD. First, we perform binary classification under both eyes-open(EO) and eyes-closed(EC) conditions using the dataset from Hospital Universiti Sains Malaysia (HUSM) (\url{https://figshare.com/articles/EEGDataNew/4244171}) \cite{mumtaz2017wavelet}. Following the same preprocessing pipeline, participants with poor signal quality are excluded, resulting in a final sample of 31 MDD patients and 21 healthy controls. The model achieve superior performance under the EC condition (Accuracy = 82.17\%$\pm$9.67\%, F1-score = 82.82\%$\pm$10.38\%), compared to the EO condition (Accuracy = 75.98\%$\pm$9.73\%, F1-score = 76.49\%$\pm$10.92\%). We further evaluate the model on the MODMA dataset using the same preprocessing pipiline, yielding 24 MDD patients and 29 healthy controls (\url{https://modma.lzu.edu.cn/data/index})\cite{cai2022multi}. Consistently, the model demonstrated robust classification performance (Accuracy = 68.59\%$\pm$7.18\%, F1-score = 70.50\%$\pm$10.98\%). As summarized in Table \ref{tab:MDD model results}, the proposed model outperform existing machine learning and deep learning approaches, with Transformer-based models ranking second. Overall, these cross-dataset validation results provide strong evidence for the robustness and generalizability of the proposed model across different datasets and experimental conditions.

\begin{table*}[!htbp]
    \caption{External validation on independent depression datasets.}
    \centering
    \begin{tabular*}{\hsize}{@{}@{\extracolsep{\fill}}ccccccc@{}}
        \toprule
        Data & Model & Accuracy(\%) & Precision(\%) & Recall(\%) & F1-score(\%) & AUC(\%) \\
        \midrule
        \multirow{16}{*}{\makecell[c]{HUSM \\ (EO)}} & KNN & 53.80$\pm$10.48 & 62.90$\pm$28.52 & 27.17$\pm$14.28 & 36.05$\pm$15.48 & 56.54$\pm$18.95\\
        & Decision tree & 53.48$\pm$11.06 & 55.20$\pm$13.84 & 54.57$\pm$12.49 & 54.03$\pm$09.47 & 53.94$\pm$11.48 \\
        & SVM & 49.13$\pm$04.31 & 49.02$\pm$05.73 & 40.43$\pm$07.19 & 44.07$\pm$06.03 & 48.31$\pm$04.43 \\
        & Random forest & 61.96$\pm$19.70 & 61.12$\pm$21.72& 76.96$\pm$23.44 & 66.75$\pm$19.43 & 60.07$\pm$24.39 \\
        & RNN & 63.37$\pm$06.91 & 58.58$\pm$22.33 & 64.13$\pm$28.22 & 59.69$\pm$21.86 & 60.78$\pm$07.92 \\
        & LSTM & 62.61$\pm$06.57 & 62.02$\pm$08.99 & 74.57$\pm$12.76 & 66.50$\pm$04.52 & 60.56$\pm$11.43 \\
        & GRU & 71.96$\pm$09.83 & 69.45$\pm$12.70 & \textbf{86.09$\pm$15.92} & 75.29$\pm$09.55 & 69.57$\pm$17.17 \\
        & Transformer & \underline{74.57$\pm$16.61} & \underline{72.65$\pm$16.78} & 80.65$\pm$21.92 & \underline{75.55$\pm$17.19} & \underline{78.01$\pm$15.73} \\
        & DDC & 59.24$\pm$07.72 & 60.84$\pm$12.64 & 65.00$\pm$23.71 & 59.76$\pm$12.84 & 56.17$\pm$09.42 \\
        & DAN & 56.52$\pm$06.66 & 50.27$\pm$28.60 & 43.70$\pm$32.78 & 43.10$\pm$25.90 & 56.20$\pm$10.42 \\
        & DANN & 59.24$\pm$06.84 & 60.17$\pm$09.49 & 68.04$\pm$20.29 & 61.49$\pm$08.52 & 57.52$\pm$10.92 \\
        & CORAL & 55.11$\pm$21.21 & 54.05$\pm$16.32 & \underline{82.17$\pm$22.02} & 64.77$\pm$17.43 & 45.15$\pm$26.28 \\
        & DCORAL & 59.24$\pm$06.97 & 59.39$\pm$09.44 & 68.91$\pm$19.28 & 62.08$\pm$08.17 & 56.50$\pm$10.50 \\
        & MoCo & 59.67$\pm$07.68 & 59.59$\pm$08.50 & 66.96$\pm$12.95 & 62.25$\pm$06.53 & 57.94$\pm$10.15 \\
        & SimCLR & 61.52$\pm$07.10 & 61.87$\pm$08.56 & 64.78$\pm$09.10 & 62.71$\pm$06.12 & 60.10$\pm$10.29 \\
        \cline{2-7}
        & \textbf{Our model} & \textbf{75.98$\pm$09.73} & \textbf{75.32$\pm$10.85} & 80.22$\pm$15.59 & \textbf{76.49$\pm$10.92} & \textbf{78.67$\pm$10.79} \\
        \midrule
        \multirow{16}{*}{\makecell[c]{HUSM \\ (EC)}} & KNN & 67.17$\pm$12.23 & 75.36$\pm$17.31 & 55.00$\pm$27.00 & 59.92$\pm$18.65 & 71.81$\pm$11.76\\
        & Decision tree & 58.15$\pm$06.36 & 58.32$\pm$06.65 & 56.74$\pm$18.03 & 56.44$\pm$10.59 & 60.13$\pm$07.54 \\
        & SVM & 45.87$\pm$08.02 & 47.18$\pm$06.44 & 58.48$\pm$17.82 & 50.89$\pm$09.60 & 51.15$\pm$10.65 \\
        & Random forest & 64.35$\pm$09.62 & 66.00$\pm$12.44 & 67.17$\pm$21.48 & 64.45$\pm$10.54 & 74.22$\pm$15.94 \\
        & RNN & 62.61$\pm$10.37 & 65.39$\pm$09.96 & 61.96$\pm$23.69 & 60.48$\pm$14.74 & 60.14$\pm$15.04 \\
        & LSTM & 64.02$\pm$06.47 & 65.40$\pm$10.60 & 68.91$\pm$18.73 & 64.92$\pm$07.36 & 59.88$\pm$12.73 \\
        & GRU & 72.61$\pm$11.40 & \underline{78.92$\pm$15.41} & 68.91$\pm$19.12 & 71.00$\pm$11.70 & 70.96$\pm$21.74 \\
        & Transformer & \underline{78.80$\pm$08.68} & 78.48$\pm$14.34 & \textbf{87.83$\pm$18.98} & \underline{80.17$\pm$09.27} & \underline{76.37$\pm$17.19} \\
        & DDC & 64.02$\pm$07.02 & 66.05$\pm$08.92 & 64.78$\pm$18.72 & 63.31$\pm$08.77 & 59.81$\pm$10.69 \\
        & DAN & 60.11$\pm$09.63 & 60.52$\pm$08.81 & 62.17$\pm$21.11 & 59.68$\pm$12.50 & 62.13$\pm$07.79 \\
        & DANN & 62.07$\pm$06.09 & 60.09$\pm$04.82 & 76.74$\pm$14.05 & 66.64$\pm$05.14 & 56.89$\pm$12.28 \\
        & CORAL & 56.74$\pm$08.31 & 61.22$\pm$16.28 & 50.65$\pm$26.53 & 51.30$\pm$14.50 & 66.43$\pm$13.98 \\
        & DCORAL & 63.91$\pm$07.90 & 66.05$\pm$11.47 & 66.52$\pm$18.87 & 63.66$\pm$10.95 & 60.39$\pm$16.22 \\
        & MoCo & 61.63$\pm$07.99 & 60.14$\pm$06.49 & 71.96$\pm$12.78 & 64.98$\pm$07.26 & 61.54$\pm$12.57 \\
        & SimCLR & 66.74$\pm$07.99 & 67.49$\pm$09.07 & 66.30$\pm$16.91 & 65.88$\pm$09.92 & 66.89$\pm$16.92 \\
        \cline{2-7}
        & \textbf{Our model} & \textbf{82.17$\pm$09.67} & \textbf{81.73$\pm$11.66} & \underline{85.22$\pm$15.05} & \textbf{82.82$\pm$10.38} & \textbf{85.80$\pm$10.53} \\
        \midrule
        \multirow{16}{*}{\makecell[c]{MODMA \\ (EC)}} & KNN & 42.83$\pm$08.42 & 37.37$\pm$12.87 & 35.00$\pm$27.56 & 34.05$\pm$18.98 & 38.01$\pm$08.96\\
        & Decision tree & 49.89$\pm$07.66 & 50.10$\pm$07.09 & 49.13$\pm$08.40 & 49.44$\pm$07.28 & 48.86$\pm$08.96 \\
        & SVM & 49.89$\pm$06.45 & 50.20$\pm$06.71 & 50.87$\pm$06.81 & 50.37$\pm$05.84 & 42.92$\pm$04.88 \\
        & Random forest & 53.04$\pm$14.44 & 52.01$\pm$17.46 & 50.65$\pm$25.48 & 49.56$\pm$20.57 & 54.19$\pm$20.70 \\
        & RNN & 56.41$\pm$03.88 & 56.43$\pm$03.92 & 64.13$\pm$20.88 & 58.09$\pm$09.37 & 54.02$\pm$05.18 \\
        & LSTM & 54.46$\pm$03.30 & 55.57$\pm$05.02 & 56.96$\pm$22.75 & 53.78$\pm$09.36 & 51.10$\pm$06.42 \\
        & GRU & 54.56$\pm$03.30 & 55.57$\pm$05.02 & 56.96$\pm$22.75 & 53.78$\pm$09.36 & 50.65$\pm$06.42 \\
        & Transformer & \underline{65.11$\pm$10.39} & \textbf{74.74$\pm$15.36} & 52.61$\pm$27.91 & 55.93$\pm$22.12 & \underline{70.09$\pm$20.20} \\
        & DDC & 51.41$\pm$03.08 & 42.74$\pm$23.22 & 66.52$\pm$44.67 & 48.56$\pm$27.53 & 50.43$\pm$04.96 \\
        & DAN & 53.26$\pm$03.69 & 44.28$\pm$24.01 & 54.78$\pm$38.63 & 46.02$\pm$25.86 & 53.21$\pm$05.84 \\
        & DANN & 55.22$\pm$04.34 & 54.92$\pm$04.46 & \textbf{69.13$\pm$23.09} & \underline{59.26$\pm$08.77} & 52.59$\pm$03.33 \\
        & CORAL & 52.72$\pm$13.96 & 52.12$\pm$16.14 & 51.09$\pm$17.82 & 51.32$\pm$16.64 & 41.89$\pm$11.81 \\
        & DCORAL & 54.23$\pm$04.11 & 50.17$\pm$18.48 & 50.00$\pm$31.03 & 47.77$\pm$19.54 & 51.00$\pm$05.18 \\
        & MoCo & 54.02$\pm$06.48 & 55.31$\pm$07.80 & 47.61$\pm$17.16 & 49.42$\pm$11.77 & 48.70$\pm$08.94 \\
        & SimCLR & 57.50$\pm$03.68 & 59.59$\pm$04.19 & 48.04$\pm$12.40 & 52.17$\pm$09.32 & 54.29$\pm$06.06 \\
        \cline{2-7}
        & \textbf{Our model} & \textbf{68.59$\pm$07.18} & \underline{71.33$\pm$11.44} & \underline{66.52$\pm$09.51} & \textbf{67.90$\pm$06.36} & \textbf{70.50$\pm$10.98} \\
        \bottomrule
    \end{tabular*}
    \label{tab:MDD model results}
\end{table*}

We further examine the microstate transition patterns captured by the model across independent MDD datasets. No significant group differences are observed under the EO condition, whereas the EC condition reveals altered bidirectional MS1$\leftrightarrow$MS3 transitions\cite{brechet2019capturing,koenig2002millisecond,milz2016functional,kim2021eeg}. These transitions have been associated with switching between internally oriented self-referential and externally oriented perceptual processing, suggesting disrupted coordination between these functional modes in MDD. Similar MS1$\leftrightarrow$MS3 alterations are observed in the MODMA dataset, providing convergent evidence that the framework captures clinically relevant disturbances in brain-state organization across independent cohorts.

\begin{figure*}[!t]
    \centering
    \includegraphics[width=\textwidth]{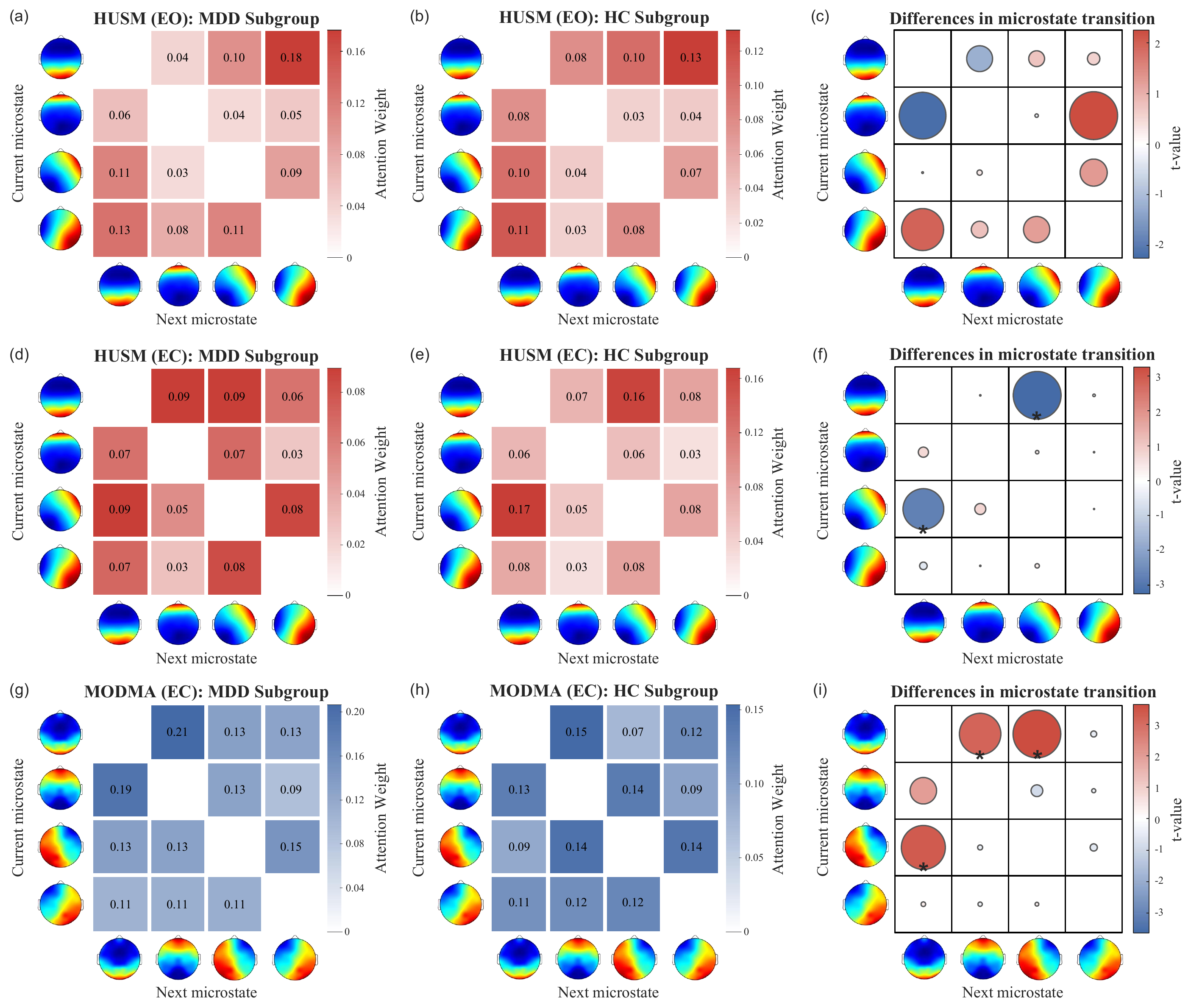}
    \caption{\textbf{Microstate transition attention weight matrices for MDD and HC in public datasets.} (a–c) HUSM under the EO conditions, (d–f) HUSM under the EC conditions, and (g–i) MODMA under the EO conditions, showing averaged ten-fold attention matrices and paired t-test results. *p$<$0.05.}
    \label{Fig: mdd supplement}
\end{figure*}

\section*{Materials and Methods}
\subsection*{Participants}
This study recruited a total of 106 adolescents with depression (age: 13–18 years; 77 females; years of education $\geq$ 5) as participants. The experiment is conducted at the Neuromodulation and Psychiatric Rehabilitation Laboratory of Shenzhen Kangning Hospital, with ethical approval from the hospital’s Research Ethics Committee (Approval No.: 2020-K021-04-1). All participants were diagnosed with MDD by psychiatrists based on DSM-5 criteria, and other psychiatric disorders were excluded through screening with the Mini International Neuropsychiatric Interview (MINI). Exclusion criteria included severe physical illness, neurological disorders, and high suicide risk. Informed consent was obtained from all participants and their families.

This study divide the depressed adolescents into two groups: depression without NSSI (DN$-$) and depression with NSSI (DN$+$). The DN$-$ group includes 39 participants (age: 15.8$\pm$2.3 years; 20 females), and the DN$+$ group includes 67 participants (age: 14.5$\pm$3.0 years; 57 females). After EEG preprocessing and exclusion of poor-quality signals, the final sample comprises 34 participants in the DN$-$ group (age: 16.0$\pm$1.39 years; 15 females) and 67 participants in the DN$+$ group (age: 14.52$\pm$1.75 years; 57 females). 

\subsection*{Data Acquisition and Preprocessing}
In this experiment, we use two types of emotion-inducing stimuli to effectively elicit negative emotions in adolescents: images of physical pain and images of social pain, with 35 images selected for each category. The physical pain images depict real injuries or skin damage, such as knife wounds, cuts, lacerations, and abrasions. The social pain images present scenarios of rejection, where an individual is visibly excluded by a group, illustrating social rejection, unfriendliness, or indifference—for example, being left out of activities, ignored, or disregarded by others.

\begin{figure*}[!t]
    \centering
    \includegraphics[width=\textwidth]{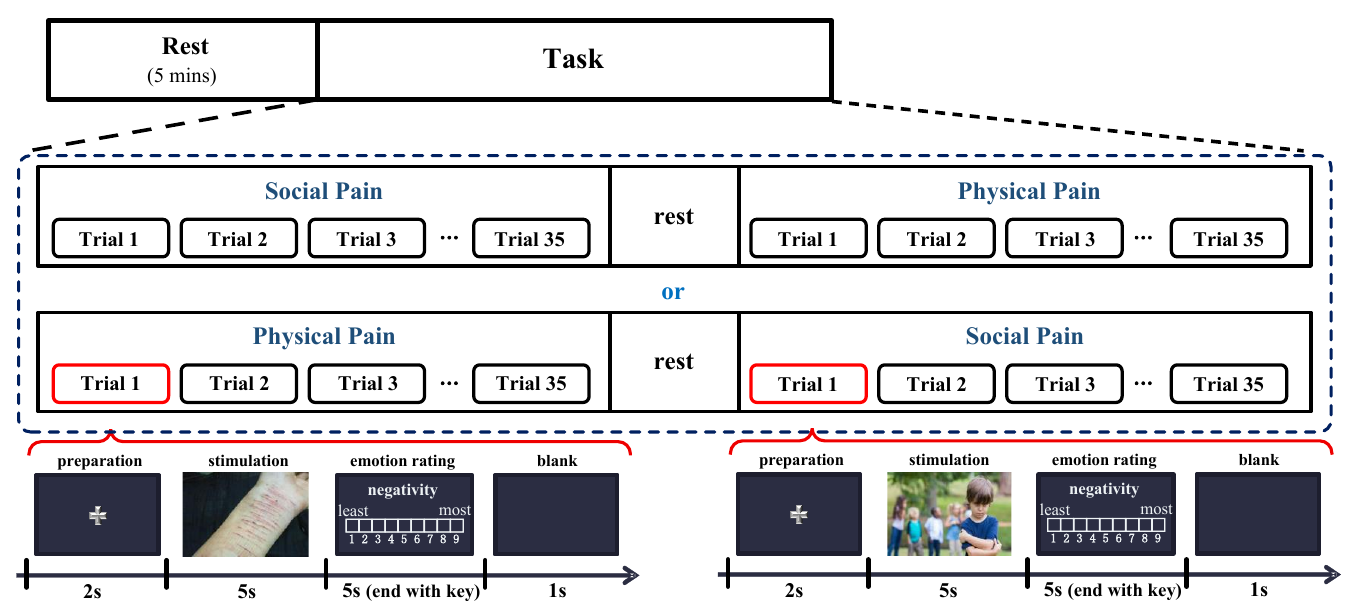}
    \caption{\textbf{Experimental paradigm.} The experiment consisted of a 5-minute resting-state session and a task session. The task session included two blocks (social pain and physical pain) presented in a randomized order. Each block comprised 35 trials, and each trial consisted of four phases: (1) Preparation phase (2 s), during which participants fixated on a central cross; (2) Stimulus phase (5 s), in which a task-relevant image is presented and participants passively viewed it; (3) Emotion rating phase (up to 5 s), during which participants rated their subjective negative emotional experience using a keyboard (1 = no negative emotion, 9 = most intense negative emotion), with the phase terminating immediately upon response; and (4) Inter-trial interval (1 s). The order of image presentation within each block was randomized across participants.}
    \label{Fig: paradigm}
\end{figure*}

We conduct the entire procedure using E-Prime 2.0 software. As shown in Fig~\ref{Fig: paradigm}, the paradigm consists of five minutes of resting-state EEG recording followed by two task conditions: physical pain and social pain, each presented randomly across 35 trials. Each trial is divided into four segments: resting state, image stimulus, emotional assessment, and a black screen. Participants first focus on a central cross for 2 s to relax, followed by a 5 s image stimulus presenting either a social exclusion or physical pain image. During the subsequent 5 s emotional assessment, participants rate the intensity of their negative emotion on a scale from 1 (no negative emotion) to 9 (strongest negative emotion) using the keyboard and confirm the rating by pressing a key. Finally, a 1 s black screen is displayed to allow relaxation between trials.

EEG signals are recorded using a 64-channel Ag/AgCl electrode cap (10–20 system) with a BrainAmp amplifier and BrainVision Recorder software at a sampling rate of 500 Hz. Offline preprocessing was performed in Letswave\cite{huang2019eeg}, an open-source toolbox running in MATLAB. Preprocessing steps include segmenting task related epochs from continuous recordings, removing unnecessary IO channels, and applying a common average reference to reduce random noise. Data are filtered with a 1–45 Hz bandpass and a 50 Hz notch filter to attenuate ocular and muscle artifacts. Independent component analysis (ICA) is used to further remove ocular artifacts. EEG segments corresponding to the image stimulation phase (5 s) are extracted, and bad channels are interpolated to ensure data quality.

\subsection*{Microstate-Based Representation of Global Brain Dynamics}
For the preprocessed EEG data, we perform microstate analysis using the open-source toolbox MST 1.0 implemented in EEGLAB to extract microstate time sequences. As illustrated in Fig~\ref{Fig: msppt step}, we first compute the Global Field Power (GFP) at each time point, defined as the standard deviation of potentials across all electrodes. The EEG topographic maps corresponding to the GFP peaks, representing moments of maximal field stability and highest signal-to-noise ratio, are then selected for subsequent microstate clustering.
\begin{equation}
    GFP(t) = \sqrt{\frac{1}N{}\sum(V_i(t)-\bar{V}(t))^{2}}
\end{equation}

Where $V_i(t)$ denotes the potential of the $i-th$ electrode at time $t$; $\bar{V}(t)$ represents the mean potential across all electrodes; and $N$ is the total number of electrodes.

Subsequently, we perform the first-stage microstate clustering at the trial level using the k-means algorithm\cite{lloyd1982least}, with the number of clusters $K=2:8$, and to ensure robustness and reproducibility, we set 2000 random initializations and 5000 iterations. The optimal number of microstate templates is determined based on high Global Explained Variance (GEV) and low Cross-Validation criterion (CV)\cite{murray2008topographic}. The optimal microstate templates obtained from all trials and participants in the previous step are subjected to a second-stage clustering using the same k-means procedure, yielding eight group-level microstate templates across participants and trials. Finally, the resulting group templates are back-fitted to the preprocessed EEG signals to derive microstate time series for each trial and each participant (each trial duration = 5 s, corresponding to 2500 sample points, hence a microstate sequence length of 2500).

\subsection*{Multidimensional Data Balancing Strategy}
Based on the collected data, we adopt a multidimensional sampling strategy to balance the dataset across gender distribution, class distribution, and trial count (DN$+$: M/F = 10/57; DN$-$: M/F = 19/15). First, 10 male and 10 female participants are randomly selected from each group (DN$+$ and DN$-$), resulting in a total of 40 participants. During model training, we employ a stratified 10-fold cross-validation approach. In each fold, one participant is selected from each subgroup—male DN$+$, female DN$+$, male DN$-$, and female DN$-$ as the target domain data for model testing (4 participants in total), while the remaining 36 participants are used as the source domain data for model training. Considering that the number of available EEG trials per participant after preprocessing varied (ranging from 19 to 35), we randomly sample 19 trials per participant to standardize the dataset. Consequently, the final model input had dimensions of $40\times19\times2500$.
This strategy effectively balanced the data distribution across gender, class, and trial dimensions, thereby mitigating potential model bias toward specific groups or conditions and enhancing the fairness and stability of the classification performance.

\subsection*{Blockwise and Randomization Strategies for Disrupting Microstate Temporal Structure}
Previous studies have reported that DN$+$ patients exhibit abnormal microstate transition patterns when exposed to social pain stimuli\cite{song2025eeg,bao2025eeg,hu2023resting}. Therefore, to enhance the model’s ability to learn NSSI related pathological microstate transition features and to reduce interference from redundant information, we apply blockwise segmentation and randomization strategies to disrupt the temporal structure of the microstate sequences before model input.

Specifically, the original microstate sequence is segmented into 50 patches using a window size of 50 and a step size of 50, as illustrated in Figure~\ref{Fig: msppt step}. Two permutation strategies are then employed to disrupt the original microstate transition patterns. 1) Strong Permutation. Randomly shuffling 40–43 patches within the sequence, effectively destroying its specific transition structure. 2) Weak Permutation. Randomly shuffling only 1–3 patches, thereby preserving most of the NSSI-related pathological microstate transition patterns while introducing mild temporal perturbations.

\subsection*{Transformer Architecture Based on Patch-Level Feature Modeling}
Compared with static microstate features (e.g., occurrence, duration, and coverage), we place greater emphasis on the dynamic transition patterns within microstate sequences, as more frequent transitions between microstates may reflect faster and more flexible information processing\cite{liu2020reliability,khanna2015microstates}. To effectively extract discriminative transition features from these sequences, we employ a patch-level Transformer model \cite{nie2022time}. Specifically, long microstate sequences are segmented into multiple patches, which helps reduce state redundancy and mitigate interference from irrelevant patterns. This design enables a hierarchical “local–global” modeling strategy, not only allowing the model to simultaneously capture fine-grained temporal dynamics within segments and the overall structure of the microstate sequence, but also enables the model to capture the temporal relationships of whole-brain activation state transitions in DN$+$ patients when exposed to social pain stimuli, thereby enhancing the representation of complex EEG dynamics.

The overall architecture of the model is illustrated in Figure~\ref{Fig: msppt step}. Inspired by the standard Transformer framework\cite{vaswani2017attention}, we first embed the original patch-level microstate sequences into a high-dimensional continuous space, allowing the model to capture the dynamic relationships and semantic structures among EEG microstates through the attention mechanism. Second, a randomly initialized CLS patch is added to the beginning of each sequence to facilitate the learning of both local and global representations of the microstate sequences. To further preserve and capture the temporal dependencies among tokens within the sequences, we introduce a dual positional encoding strategy to enhance the model’s ability to represent temporal structure. Specifically, the first positional encoding is applied before patch segmentation to retain token-level temporal dependencies, while the second positional encoding is applied after adding the CLS patch to model patch-level temporal relationships, thereby reinforcing temporal representation at both local and global levels. Finally, the processed sequences are fed into a three-layer Patch-Transformer, where they are transformed into feature vectors, and the weighted attention matrices from each layer are retained for subsequent analysis.

\subsection*{Multi-loss Joint Optimization}
Although conventional Transformer models are easy to train, they often produce low-quality features, limiting their effectiveness for complex tasks. In contrast, ordered deep modeling of microstate sequences has shown great potential in generating high-quality features from time-series data. To enhance the model’s robustness and accuracy in distinguishing between the two participant groups—particularly in identifying the pathological microstate transition patterns of DN$+$ patients—we incorporate consistency learning and contrastive learning modules. These components aim to strengthen the model’s ability to capture core pathological features through mechanisms that enforce both consistency and discriminative representation learning.

\begin{itemize}[leftmargin=*, labelsep=4pt]
    \item Disease Specific Loss. To effectively capture the microstate transition differences between DN$+$ and DN$-$ participants, we design a disease specific loss. This loss plays a crucial role in enhancing the model’s discriminative capability, ensuring that features unique to each condition are accurately identified. The disease-specific loss function is defined as follows:
    \begin{equation}
        \mathcal{L}_{Dis}=-\frac{1}{N}\sum_{i=1}^{N}(y_ilog(\hat{y_i})-(1-y_i)log(1-\hat{y_i}))
    \end{equation}
    Where $y_i$ denotes the true label, $\hat{y_i}$ represents the predicted label, and $N$ is the number of samples.
    \item Domain Adversarial Loss. To mitigate the influence of individual differences on the model’s ability to distinguish between participant groups, we incorporate a domain adversarial loss. This loss ensures that the model learns domain invariant features, thereby enhancing its generalization capability in cross-domain tasks. The domain-adversarial loss function is defined as follows:
    \begin{equation}  
    \begin{aligned}
        \mathcal{L}_{Adv}
        =
        -\frac{1}{2}
        \sum_{x_s \in \mathcal{D}_s}
        \log D\!\left(R_{\lambda}(F(x_s))\right)\\
        -\frac{1}{2}
        \sum_{x_t \in \mathcal{D}_t}
        \log\left(1-D\!\left(R_{\lambda}(F(x_t))\right)\right)
    \end{aligned}
    \end{equation}
    where $x_s$ and $x_t$ denote samples drawn from the source domain $\mathcal{D}_s$ and target domain $\mathcal{D}_t$, respectively. $F(\cdot)$ denotes the feature extractor, $D(\cdot)$ represents the domain discriminator, $R_{\lambda}(\cdot)$ is the gradient reversal layer, and $\lambda$ denotes the gradient reversal coefficient.
    \item Consistency Learning Loss. Previous studies have shown that DN$+$ participants exhibit abnormal microstate transition patterns in response to pain-related emotional stimuli\cite{song2025eeg,bao2025eeg,hu2023resting}. To ensure that microstate sequences with pathological transition features are accurately identified, we design a consistency learning module. In this module, a single-layer unidirectional LSTM is first applied to both the original microstate sequences and the strongly permuted microstate sequences. The final hidden state of each LSTM serves as a context vector, effectively summarizing the patch sequence. During the pseudo-label training phase, the model performs consistency learning by distinguishing the context vectors of the original sequences (label = 1) from those of the strongly permuted sequences (label = 0). The time-and-channel patch consistency loss is then defined based on binary cross-entropy. The consistency learning loss function is formulated as follows:
    \begin{equation}
        \mathcal{L}_{CS}=-\frac{1}{N}\sum_{i=1}^{N}
        \left(y_i\log(\hat{y_i})+(1-y_i)\log(1-\hat{y_i})\right)
    \end{equation}
    Where $y_i$ denotes the pseudo label, $\hat{y_i}$ represents the predicted label, and $N$ is the number of samples.
    \item Contrastive Learning Loss. To further amplify subtle differences between microstate sequences of DN$+$ and DN$-$ participants and strengthen classification boundaries, we employ the InfoNCE loss to construct a more precise feature discrimination mechanism\cite{oord2018representation}. Similarly, a single-layer unidirectional LSTM is first applied to the original microstate sequences, strongly permuted sequences, and weakly permuted sequences, with the final hidden state of each LSTM serving as a context vector. In this setup, the original sequence acts as the anchor, the weakly permuted sequence as the positive pair, and the strongly permuted sequence as the negative pair. The contrastive loss over temporal sequences is then defined as follows:
    \begin{equation}
        \mathcal{L}_{CT}=-log(\frac{exp(sim(c_{o},c_{w})/\tau)}{exp(sim(c_{o},c_{w})/\tau)+exp(sim(c_{o},c_{s})/\tau)})
    \end{equation}
    Where $c_{o}$,$c_{w}$ and $c_{s}$ denote the context vector of the original sequence, weak permutation, and strong permutation sequences, $\tau$ is the temperature parameter, and $sim$ represents the cosine similarity.
    \item Total Loss Function. The overall loss for multi-loss joint optimization is defined as the direct sum of all individual loss components. In summary, the classification (disease-specific) loss establishes the model’s basic ability to distinguish between participant groups, the domain-adversarial loss enhances the model’s generalization capability, the consistency loss ensures the stability of learning meaningful features, and the contrastive loss improves the precision in capturing subtle differences.
    \begin{equation}
        \mathcal{L}_{Total} = \alpha \mathcal{L}_{Dis}+\beta\mathcal{L}_{Adv}+\delta\mathcal{L}_{CS}+\theta\mathcal{L}_{CT}
    \end{equation}
    where $\alpha$, $\beta$, $\delta$ and $\theta$ are weighting factors for the corresponding loss terms.
\end{itemize}

The overall loss function ensures that the model learns robust, multi-dimensional EEG feature representations that are invariant to signal noise, sensitive to gender differences, adaptable across domains, and effective in classifying NSSI behaviors. By jointly optimizing these objectives, the model is better equipped to handle the complexities involved in EEG signal analysis for studies on depression and NSSI.

\subsection*{Statistical Analysis}
\textcolor{black}{For the cosine similarity between the learned neural representations and stimulus image features under different pain conditions, statistical comparisons between the two groups are performed using two-way mixed ANOVA. Differences in microstate transition attention weights between disease and sex subgroups are evaluated using paired t-tests on the attention weight matrices obtained from each cross-validation fold. Behavioral analyses are conducted using correlation and mediation analyses. For the temporal validity analysis, independent two-sample t-tests are performed to compare microstate features between the two groups. Significance is indicated as follows in the figures: $*p\leq0.05$; $**p\leq0.01$; $***p\leq0.001$.}

\section*{Acknowledgments}
\textcolor{black}{This work was supported by the National Natural Science Foundation of China under Grant 62522608, the Guangdong Provincial Project (Grant No. 2024TQ08X330), Brain Science and Brain-like Intelligence Technology - National Science and Technology Major Project (2021ZD0200500), Shenzhen-Hong Kong Institute of Brain Science-Shenzhen Fundamental Research Institutions (2023SHIBS0003), the Shenzhen Science and Technology Program (No. JCYJ20241202124222027 and JCYJ20241202124209011), and the Key Research and Development Program of Hunan Province (2025QK3008).}

\section*{Author Contributions} 
\textbf{Ying Xu:} Conceptualization, Methodology, Formal analysis, Investigation, Writing – original draft. \textbf{Xiaojun Liang:} Methodology, Validation, Investigation. \textbf{Li Zhang:} Writing – review \& editing. \textbf{Yixuan Yuan:} Investigation, Writing – review \& editing. \textbf{Gan Huang:} Data curation, Visualization.
\textbf{Yongjie Zhou:} Resources, Investigation, Writing – review \& editing. \textbf{Zhen Liang:} Conceptualization, Methodology, Supervision, Project administration, Funding acquisition, Writing – review \& editing.

\section*{Conflicts of Interest}
The authors declare that there is no conflict of interest regarding the publication of this article.

\section*{Data Availability}
The data that support the findings of this study are available from the corresponding author upon reasonable request. 


\printbibliography

\end{document}